# Continuous Dice Coefficient: a Method for Evaluating Probabilistic Segmentations

Reuben R Shamir, Yuval Duchin, Jinyoung Kim, Guillermo Sapiro, and Noam Harel

**Abstract— Objective:** Overlapping measures are often utilized to quantify the similarity between two binary regions. However, modern segmentation algorithms output a probability or confidence map with continuous values in the zero-to-one interval. Moreover, these binary overlapping measures are biased to structure's size. Addressing these challenges is the objective of this work. *Methods:* We extend the definition of the classical Dice coefficient (*DC*) overlap to facilitate the direct comparison of a ground truth binary image with a probabilistic map. We call the extended method *continuous Dice coefficient* (*cDC*) and show that 1) *cDC* ≤1 and *cDC* = 1 if-and-only-if the structures' overlap is complete, and; 2) *cDC* is monotonically decreasing with the amount of overlap. We compare the classical *DC* and the *cDC* in a simulation of partial volume effects that incorporates segmentations of common targets for deep-brain-stimulation. Lastly, we investigate the *cDC* for an automatic segmentation of the subthalamic-nucleus. *Results:* Partial volume effect simulation on thalamus (large structure) resulted with *DC* and *cDC* averages (SD) of 0.98 (0.006) and 0.99 (0.001), respectively. For subthalamic-nucleus (small structure) *DC* and *cDC* were 0.86 (0.025) and 0.97 (0.006), respectively. The *DC* and *cDC* for automatic STN segmentation were 0.66 and 0.80, respectively. *Conclusion:* The *cDC* is well defined for probabilistic segmentation, less biased to structure's size and more robust to partial volume effects in comparison to *DC*. *Significance:* The proposed method facilitates a better evaluation of segmentation algorithms. As a better measurement tool, it opens the door for the development of better segmentation methods.

## I. INTRODUCTION

Accurate segmentation of anatomical or pathological structures (regions) on medical images facilitates effective and safer surgical planning [1-2] and quantitative monitoring of disease progression [3]. To evaluate a segmentation method, the computed and ground truth segmented regions are compared. Overlap measures, such as the Dice coefficient (*DC*), which operates on binary data, are often computed [4]–[7].

Submitted on: February 03, 2016. Work partially supported by NIH grants R01-NS085188, P41-EB015894, P30-076408, and Surgical Information Sciences, Inc.

Reuben R Shamir and Yuval Duchin are with Surgical Information Sciences, Minneapolis, MN, USA (correspondence e-mail: shamir.ruby@gmail.com). Yuval Duchin and Noam Harel are with the University of Minnesota, Twin Cities, MN, USA. Jinyoung Kim and Guillermo Sapiro are with Duke University, Durham, NC, USA.

Crum *et al.* [4] generalized these (binary) overlap measures to measure the total overlap of ensembles of labels defined on multiple test images and to account for fractional labels using fuzzy set theory. Rohlfing *et al.* [5] showed that the Dice coefficient is directly related to structure's size: the smaller the structure the lower the Dice coefficient (given a fixed resolution). Therefore, Dice is a difficult measure for comparing methods tested on different structures, complicating the design of an effective segmentation approach.

Many modern automatic and semi-automatic segmentation methods output a probabilistic (or confidence) map, that is, an image with real values in the [0, 1] interval. In this case, the common overlap measures, such as *DC*, do not apply and the probabilistic map needs to be converted (usually with a threshold) into a binary image beforehand. However, this conversion varies the original segmentation and does not necessary represents its actual original quality. Zou *et al.* [8] have addressed this issue and suggested a numerical integration method to compute the *DC* for a probabilistic map under the assumption of a uniform prior distribution in [0, 1] of a threshold parameter. As the authors demonstrate, drawing the threshold parameter from another distribution may change the *DC* value substantially [8]. The optimal threshold distribution or how to revise the method for other distributions is unclear. Moreover, the method requires a numerical integration method that may be time-consuming, hard to optimize, and provides only an approximate solution.

In this note we introduce a closed-form method that extends the definition of Dice coefficient and that does not require a threshold on the probabilistic segmentation map. The proposed extended version, here denoted as *continuous Dice coefficient* (*cDC*), addresses the above limitations, the size-dependency of the classical *DC* and incorporates the probabilistic nature of the segmentation.

## II. CONTINUOUS DICE COEFFICIENT

The classical Dice coefficient is defined as

$$DC := \frac{2|A \cap B|}{|A| + |B|}. \tag{1}$$

Here *A* is a set representing the ground-truth and *B* represents the computed segmentation. Both images (sets) are binary with values '0' or '1' at each of their voxels (or pixels in the 2D case). These values are denoted here as $a_i$ and $b_i$, respectively. In this case, Equation (1) can be computed as follows:

$$|A \cap B| = \sum_i a_i b_i, \tag{2}$$

$$|A| = \sum_i a_i, \tag{3}$$

$$\text{and } |B| = \sum_i b_i. \tag{4}$$

Many segmentation methods output a probabilistic map, where voxels are associated with a real value ($b_i \in [0,1]$). In this case,[1] $\sum_i a_i b_i < \sum_i a_i$ (we assume that not all $b_i$ are 1) and therefore $DC < 1$ also when $A$ and $B$ completely overlap (i.e., $\forall b_i > 0, a_i = 1$ and $\forall b_i = 0, a_i = 0$). Moreover, the above formulas in the classical $DC$ ignore the provided confidence values, which are critical to evaluate the quality of the segmentation. This confidence is often low at segment boundaries (e.g., due to resolution and partial volume effects), thereby having a larger effect on small segments.[2]

In this note we address these issues and suggest extending the definition of the $DC$ to enable the direct comparison of continuous measures with the ground truth segmentation. Specifically, we weight $|A|$ such that the $DC$ value becomes 1 at complete overlap (as defined above) and define the *continuous Dice coefficient* ($cDC$) as

$$cDC := \frac{2|A \cap B|}{c|A| + |B|}, \tag{5}$$

where $c$ is defined as the mean value of $B$ over the voxels where both $A$ and $B$ are positive and can be computed as

$$c = \frac{\sum_i a_i b_i}{\sum_i a_i sign(b_i)}, \tag{6}$$

where $sign(x)$ is defined as

$$sign(x) = \begin{cases} 1 \ if \ x > 0 \\ 0 \ if \ x = 0 \\ -1 \ if \ x < 0 \end{cases}. \tag{7}$$

If $\sum_i a_i sign(b_i) = 0$ (no overlap between $A$ and $B$), we arbitrary define $c = 1$. In this case, $cDC$ will be zero since $\sum_i a_i b_i = 0$. Moreover, note that when $b_i \in \{0, 1\}$ (i.e., a binary value) $c = 1$ and $cDC = DC$. Therefore, $cDC$ is a consistent extension of the $DC$ to the more general case of real values in the [0, 1] interval. The proposed $cDC$ has a number of key properties that we present next.

[1] For simplicity of the presentation we consider the ground truth $A$ a binary mask, though the proposed measurement can be extended to probabilistic ground truth as well.

[2] For example, with a 1mm standard MRI resolution, a region of 5mm width (common in surgical targets) has about 30% of voxels in the boundary, suffering from resolution and partial volume effects.

*Property I: $cDC \leq 1$ and $cDC = 1$ iff overlap is complete*

Proof: When $A$ and $B$ completely overlap (e.g., $\forall b_i > 0, a_i = 1$ and $\forall b_i = 0, a_i = 0$), then $|A \cap B| = \sum_i a_i b_i = \sum_i b_i = |B|$. Morever,

$|A \cap B| = \sum_i a_i b_i = \sum_i a_i b_i \frac{\sum_i a_i}{\sum_i a_i} = \frac{\sum_i a_i b_i}{\sum_i a_i sign(b_i)} \sum_i a_i = c|A|.$

Therefore, $2|A \cap B| = c|A| + |B|$ and $cDC = 1$.

When $A$ and $B$ partially or not overlap at all (e.g., $\exists a_i = 1$ and $b_i = 0$ or $\exists b_i > 0$ and $a_i = 0$), $|A \cap B| = \sum_i a_i b_i < \sum_i b_i = |B|$ or, $\sum_i a_i > \sum_i a_i sign(b_i)$ and then $|A \cap B| = \sum_i a_i b_i = \sum_i a_i b_i \frac{\sum_i a_i}{\sum_i a_i} < \frac{\sum_i a_i b_i}{\sum_i a_i sign(b_i)} \sum_i a_i = c|A|.$ Therefore, $2|A \cap B| < c|A| + |B|$ and $cDC < 1$.

*Property II: $cDC$ is monotonically decreasing with the amount of overlap*

Proof: Let's assume that $B$ and $D$ are two probabilistic maps of the "same size" (i.e., $|B| = |D|$) and that the segmented structure is represented by the ground-truth binary image $A$. Now, if the overlap (as defined above, where both are non-zero) of $A$ and $B$ is smaller than the overlap of $A$ and $D$ (e.g., $|\{i: a_i = 1 \ and \ b_i > 0\}| < |\{i: a_i = 1 \ and \ d_i > 0\}|$), then we have that $\sum_i a_i sign(b_i) < \sum_i a_i sign(d_i)$. Since $|B| = |D|$ and $|A|$ and 2 remain the same in the computation of the $cDC$ (Equation (5)), it is enough to compare $\frac{|A \cap B|}{c(A,B)}$ with $\frac{|A \cap D|}{c(A,D)}$.

$$\frac{|A \cap B|}{c(A,B)} = \frac{\sum_i a_i b_i}{\frac{\sum_i a_i b_i}{\sum_i a_i sign(b_i)}} = \sum_i a_i sign(b_i) < \sum_i a_i sign(d_i) = \frac{|A \cap D|}{c(A,D)}$$

Therefore $cDC(A,B) < cDC(A,D)$.

Note that if B and D are not of the same size, a method for comparing the overlaps is undefined. Various methods for overlap comparison can be crafted to demonstrate Property II (in the most trivial solution we can define it as $cDC$).

TABLE I

CONTINUOUS DICE COEFFICIENT MATLAB IMPLEMENTATION

```
function  = continuous_dice_coefficient (A,B)

size_of_A_intersect_B = sum(A(:).*B(:));
size_of_A = sum(A(:));
size_of_B = sum(B(:));

if (size_of_A_intersect_B > 0)
    c = sum(A(:).*B(:))/sum(A(:).*sign(B(:)));
else
    c = 1;
end

cDC=(2*size_of_A_intersect_B) / (c*size_of_A + size_of_B);
```

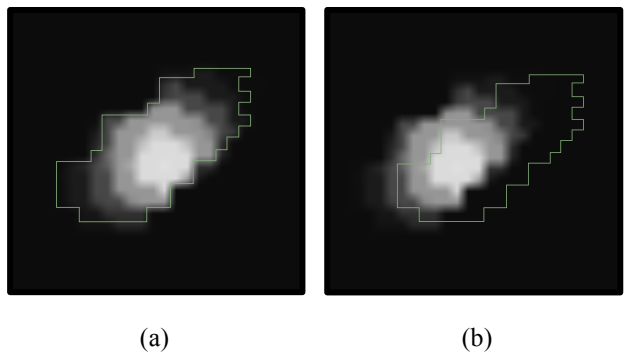

(a)                                  (b)

Fig. 1. Empirical illustration of the proposed *cDC*. (a) A probabilistic map was simulated with a Gaussian distribution over a manually segmented image of the subthalamic nucleus (green line marks its boundaries at a selected plane). (b) Then, the probabilistic map was shifted to simulate a simple segmentation error (2mm in this example). The proposed *cDC* was computed under the various translation errors to empirically confirm the properties of *cDC*. Moreover, a random-direction half-voxel translation simulated partial volume effect to compare the *cDC* with *DC*.

## III. COMPARISON WITH DICE COEFFICIENT

We implemented and evaluated the presented continuous dice coefficient with MATLAB (The MathWorks Inc., MA, USA). Our implementation for the *cDC* is presented in Table 1. Properties I and II were empirically confirmed by simulations on manipulated clinical data (Fig. 1).

To compare the *cDC* with *DC* we manually segmented the right subthalamic nucleus (STN), globus pallidus (GP), and thalamus on a high-field high-resolution 7T MRI head image of a Parkinsons' disease patient [9]. Then, we copied the binary segmentation and replaced its '1's with a simulated probabilistic segmentation map with a Gaussian distribution with respect to the center of the structures (Fig. 1a). This type of confidence distribution is expected due to resolution and partial volume effects. Then, we translated the simulated probabilistic image at a random direction 0.25mm (half-voxel) to simulate partial volume effects and measured the *cDC*. In addition, we translated a copy of the original binary segmentation using the same transformation and measured the *DC*. We repeated this process for 20 random translations.

Next we evaluate the proposed *cDC* on an actual computational segmentation method and compare it with the classical *DC*. To this end, we automatically computed the segmentation probabilistic map of the right STN on a standard clinical MRI of a Parkinson's disease (PD) patient. We have a database of 46 PD patients that incorporates co-registered standard clinical MRI, high-field 7T MRI and a segmentation of anatomies in the basal ganglia for each patient. We aligned the 16 most similar clinical images with that of the new (out-of-database) patient to create an initial guess and then applied machine learning algorithms that eventually output a probabilistic map regarding the location and shape of the right STN. We refer the reader to [10] for more details about the method.

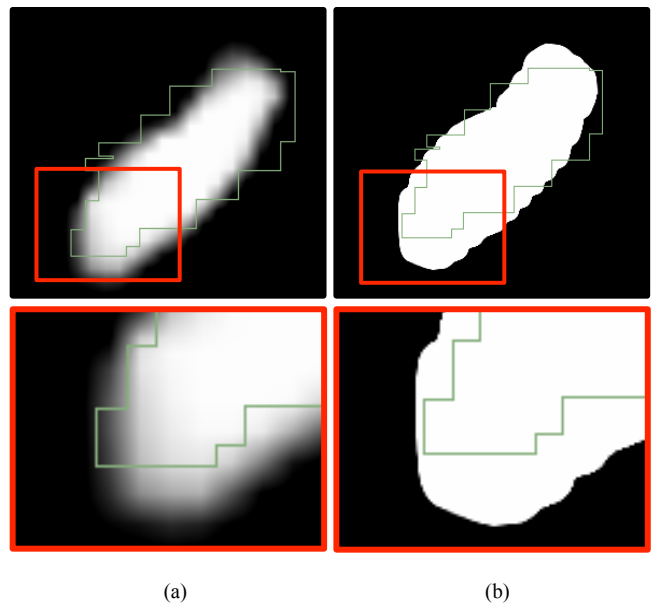

(a)                                  (b)

*Fig. 2.* A comparison of the proposed continuous and classical binary Dice coefficients. (a) Ground truth segmentation of the STN (green) and the computed probabilistic map that was used for the computation of the *cDC*. (b) A binary image produced by applying a threshold of 0.1 on the probabilistic map. It was used for the computation of the *DC*. In this example the *cDC* associates low weights for the false positive errors at the bottom of the STN according to their probability. The classical *DC* fails to incorporate the probabilistic information and counts all errors/inaccuracies equally.

The ground truth of the right STN segmentation was extracted from a co-registered 7T MR image of the same patient (Fig. 2). A threshold of 0.1 was applied to the segmentation probabilistic map to convert it into a binary image and compute the *DC*. This threshold was selected to achieve maximal *DC* value. The *cDC* was computed directly on the segmentation probabilistic map and compared to the *DC*.

## IV. RESULTS

Properties I and II were empirically confirmed by simulations on manipulated clinical data. Fig. 3 shows that *cDC* is less biased and more robust to partial volume effects in comparison to the classical *DC*: compare STN averages (SD) of 0.86 (0.025) and 0.97 (0.006) of *DC* and *cDC*, respectively. Another interesting observation is that the *DC* is highly related to structure size: the smaller the structure, the lower the *DC* value (Fig. 3).

The *DC* and *cDC* values between automatically segmented and ground truth right STNs were 0.66 and 0.80, respectively. Fig. 2 demonstrates that the proposed *cDC* accounts for the probability values of the segmentation algorithm. The *DC* is missing these values and all errors are counted equally.

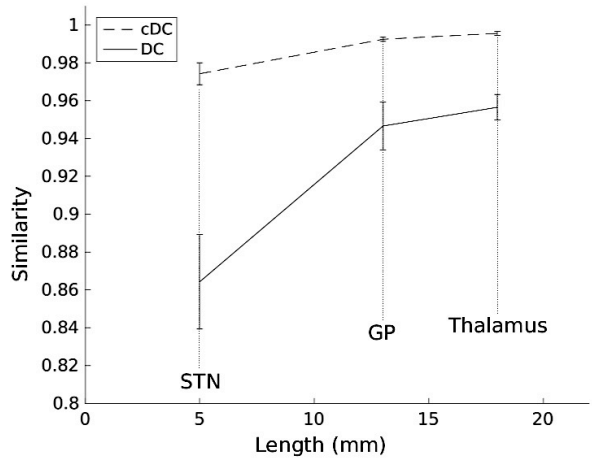

Fig. 3. Partial volume effect on continuous- and classical- Dice coefficients (*cDC* and *DC*, respectively). Half-voxel translations at random directions (linear interpolation) simulate partial volume effect. Average similarity values are presented along with the standard deviation (error bars). The shorter the structure, the lower the *DC* and *cDC* and their variance increases. However, the effect on *DC* is much more prominent: compare STN averages (SD) 0.86 (0.025) and 0.97 (0.006) of *DC* and *cDC*, respectively.

## V. DISCUSSION

Our results show that the proposed continuous Dice coefficient is an appropriate measure of similarity between a binary ground-truth and a computed probabilistic segmented image. Current overlap measures, such as the Dice coefficient, require applying a threshold on the probabilistic map. This process counts all potential errors as even and thereby ignoring issues such as image resolution and partial volume effects. The proposed continuous Dice coefficient accounts for partial errors and/or low confidence pixels/voxels. Our results show that it was less biased and more robust to partial volume effect and structure's size in comparison to the classical Dice coefficient.

Small structures have relatively large number of voxels on their surface with respect to their inner zones. It is expected that the probabilistic map will have lower values on these boundaries. When the Dice coefficient is used, many of the voxels around the boundaries are considered as error. The STN for example is about 4mmx6mmx8mm. Considering the standard clinical image resolution of ~1mm$^3$, about one third of its voxels are on the boundaries. This may explain the low Dice coefficient value of 0.66 that was observed in our experiment. However, the proposed continuous Dice coefficient counts only the partial errors and is less biased by structure's size (Fig. 3). Therefore, it resulted with a higher value of 0.8, and much reduced variance. Our simulations suggest that the *cDC* better reflects the actual quality of the segmentation.

## VI. CONCLUSION

We extended the commonly used Dice coefficient measure to enable the direct comparison of a computed probabilistic map with ground truth segmentations. We have shown that the proposed continuous Dice coefficient satisfies desired properties and that it is less biased and more robust in comparison to the classical Dice coefficient. The proposed continuous Dice coefficient weights the segmentation errors according to their confidence/probability, as opposed to the classical Dice coefficient that rates them all the same. We expect that this new measure will assist in studies on probabilistic segmentation methods and with the design and analysis of new techniques.